# HANet: A hierarchical attention network for change detection with bi-temporal very-high-resolution remote sensing images[1]

Chengxi Han, *Student Member, IEEE*, Chen Wu, *Member, IEEE*, Haonan Guo, *Student Member, IEEE, Meiqi Hu, Student Member*, *IEEE*, Hongruixuan Chen, *Student Member, IEEE*

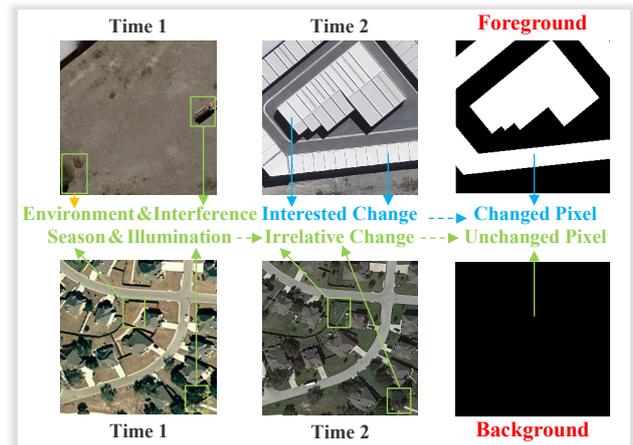

**Fig. 1.** The main data categories for VHR image CD include foreground and background images. The foreground image is an interesting change containing a region of interest, such as a building. The background image is an irrelevant change containing seasonal, illumination, environmental, and other interference changes.

*Abstract*—Benefiting from the developments in deep learning technology, deep learning-based algorithms employing automatic feature extraction have achieved remarkable performance on the change detection (CD) task. However, the performance of existing deep learning-based CD methods is hindered by the imbalance between changed and unchanged pixels. To tackle this problem, a progressive foreground-balanced sampling (PFBS) strategy on the basis of not adding change information is proposed to help the model accurately learn the features of the changed pixels during the early training process and thereby improve detection performance. Furthermore, we design a discriminative Siamese network, Hierarchical Attention Network (HANet), which can integrate multi-scale features and refine detailed features. The main part of HANet is the HAN module, which is a lightweight and effective self-attention mechanism. Extensive experiments and ablation studies on two CD datasets with extremely unbalanced labels validate the effectiveness and efficiency of the proposed method. Our model is available at https://github.com/ChengxiHAN/HANet-CD.

## I. INTRODUCTION

CHANGE detection (CD) is the process of identifying differences in the state of an object or phenomenon by observing it at different times [1]. The goal of binary CD is to assign binary labels (i.e., change or no change) to every pixel in a region [1]. Very-high-resolution (VHR) remote sensing (RS) imagery CD is one of the fundamental topics in the field of RS image interpretation, and has a wide range of applications, including land use land cover analysis [2], urban extension studies [3], environmental monitoring [4], and disaster assessment [5]. Notably, the feature extraction of temporal RS images will be negatively influenced due to differences in the light illumination, contrast, quality, resolution, and noise of RS images in the

[1] This work was supported in part by the National Key Research and Development Program of China under Grant 2022YFB3903300 and 2022YFB3903405, and in part by the National Natural Science Foundation of China under Grant T2122014, 61971317, 62225113 and 42230108. (*Corresponding author: Chen Wu.*)

Chengxi Han, Chen Wu, Haonan Guo and Meiqi Hu are with the State Key Laboratory of Information Engineering in Surveying, Mapping and Remote Sensing, Wuhan University, Wuhan 430079, China (e-mail: chengxihan@whu.edu.cn; chen.wu@whu.edu.cn; guohnwhu@163.com; meiqi.hu@whu.edu.cn).

Hongruixuan Chen is with the Graduate School of Frontier Sciences, The University of Tokyo, Chiba 277-8561, Japan (e-mail: Qschrx@gmail.com).

same region at different times, which makes RS image change detection a challenging field of research.

CD is one of the hot topics in the field of RS, and can be broadly divided into traditional methods and deep learning methods. Researchers have devised a large number of traditional and deep learning methods of CD. Among these, the traditional methods depend primarily on original image information and handcrafted features. Bruzzone [6] proposes the change vector analysis (CVA) technique to calculate the intensity and direction of two or more types of change. Nielsen [7] proposes the multivariate alteration detection (MAD) method to maximize the variance of the transformed variable; this approach is insensitive to affine transformations. In [8], principal component analysis (PCA) is proposed to convert a differential or stacked image into a new feature space, making this approach a feature transformation method. Wu [9] proposes the use of slow feature analysis (SFA) to extract the most invariant component from a multi-temporal remote sensing image and convert the original image into a new feature space, in which the changed pixels are highlighted while the unchanged pixels are suppressed.

Subsequently, the method of post-classification comparison (PCC) was developed. Through sample selection and the feature classification of two RS images, the detailed change information of features can be obtained by comparing the classification map [10]. However, the classification comparison method often requires a large amount of training data; moreover, the detection accuracy depends entirely on the initial classification accuracy.

Although these traditional CD methods have achieved good detection results in their respective application scenarios, due to the limitations of these traditional manually designed feature models, they only use the spectral information of multi-temporal image data. Thus, traditional methods are mainly utilized on medium- and low-resolution remote sensing images. In comparison, VHR images contain many more spatial details; as a result, traditional methods will lead to internal fragmentation phenomena appearing in the changed area and salt and pepper noise in the unchanged area.

The generalization and accuracy of the traditional methods for big-data CD tend to be limited since they depend primarily on original image information and handcrafted features. For this reason, researchers have worked to develop automatic change detection methods based on deep learning, which can achieve better performance.

Deep learning for change detection is one of the current hottest topics in the field of RS, as it enables the extraction of representative deep features. In [11], three fully convolutional neural network architectures (FC-EF, FC-Siam-conc, and FC-Siam-diff) are proposed. Two of these are based on Siamese networks; this represents the first time that a skip connection has been added on the basis of a Siamese network. Moreover, the attention mechanism [12] enables the model to focus on, and to fully learn and absorb, important information, which is then introduced to the CD; some works employing this approach include STANet [13], SNUNet [14], and MSPSNet [15]. Over the years, the transformer mechanism [16] has been very popular in the field of natural language processing, and was subsequently introduced into the computer vision and CD contexts. Examples include BIT [17], Change Former [18], and RSP-BIT [19]. Notably, while these deep learning methods all achieve good results in their own scenarios, they do not focus on the problem of sample imbalance.

As shown in Fig.1, the goal of CD is to find the changed pixels (white) and unchanged pixels (black). In general, there are two categories used in VHR image CD, namely foreground and background images. The foreground is a relevant change containing a region of interest such as a building. The background is an irrelevant change due to seasonal, illumination, environmental, and other interference changes.

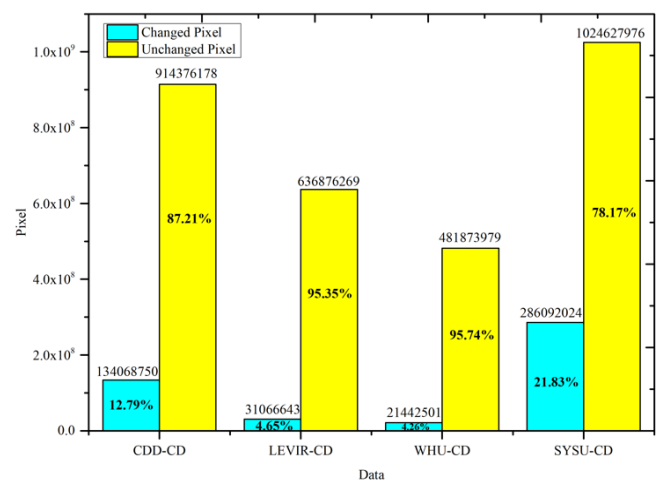

**Fig.2.** Statistics related to the number of changed and unchanged pixels (CDD-CD, LEVIR-CD, WHU-CD, and SYSU-CD).

At the same time, as shown in Fig.2, we compile statistics related to the number of changed and unchanged pixels (CDD-CD [20], LEVIR-CD [13], WHU-CD [21], and SYSU-CD [22]). From Fig.2, it can be observed that the percentage of changed pixels is extremely low; we refer to this as an **extremely unbalanced binary CD**.

However, most existing methods tend to ignore the imbalance problem and learn the features from the data directly, which leads to the model cannot sufficiently learn the features of the data. A few existing methods have attempted to solve the imbalance problem by designing new loss functions (e.g., dice loss and weighted cross-entropy loss). In addition, some attention-based methods struggle to relate to long-range concepts in space-time, while the computational complexity of some self-attention-based methods is unacceptably high. Thus, it is necessary to develop more efficient and lightweight attention mechanisms. And also some methods synthesize CD samples like CDNet+IAug [23] and transfer a pre-trained model like SaDL [24] to make it more robust, In fact, it is the addition of changing pixels to improve the effectiveness of the model.

To better solve these problems, we propose a set of Progressive Foreground-Balanced Sampling (**PFBS**) approaches on the basis of not adding change information to help the model accurately learn the features of the foreground image during the early stages of the training process, and design a more discriminative Siamese network, Hierarchical Attention Network (**HANet**), to integrate multi-scale features and refine detailed features. The main contributions of this work can be summarized as follows:

1) An original Progressive Foreground-Balanced Sampling (PFBS) strategy on the basis of not adding change information is put forward to deal with the data-imbalance challenge of binary change detection without additional computation cost. The progressive policy firstly studies the minority foreground samples in a centralized manner, empowering the network to learn the most significant characteristic of change features.
2) A discriminative Siamese Hierarchical Attention Network (HANet) is tailored to integrate multi-scale features and refine detailed spatial and temporal change features, where a lightweight and effective HAN module is capable of capturing long-term dependencies separately from the column and row dimensions.
3) Extensive experiments and ablation studies on two extremely unbalanced binary CD datasets validate the effectiveness and efficiency of the proposed method. We open-source the code and hope to contribute to the field of CD research.

The rest of this paper is organized as follows. Section II describes the previous works related to deep learning-based methods and the attention mechanism in the CD context. In Section III, the structure of the proposed HANet is illustrated in detail. Section IV reports the experimental results and ablation study. Finally, conclusions are provided in Section V.

## II. RELATED WORK

In this section, we briefly introduce the deep learning-based method and the attention mechanism of CD.

### A. Deep Learning-based CD Methods

Deep convolutional neural networks (CNNs) can extract hierarchical features. Among them, high-level features contain abstract semantic information, while low-level features contain rich spatial details. Therefore, high-level semantic information is crucial to understanding scenes with complex backgrounds. This has led to the rise of RS interpretation methods based on deep learning, such as object detection [25], [26], classification [27], semantic segmentation [28], [29], and CD [30], [31].

Performing CD on RS images requires pixel-level prediction, and the fully convolutional neural network [32] is mainly used for intensive prediction tasks. Therefore, CD methods based on deep learning mainly use structures similar to the fully convolutional network, then generate difference maps or change vectors by comparing features of different depths. In [33], a deep Siamese convolutional neural network was used to capture similar features of bi-temporal RS images, after which the K-nearest neighbor method was used to cluster similar pixels in the feature map to obtain the regions of variation.

These methods are a combination of deep neural networks and machine learning algorithms. Notably, although these methods have achieved better results, the whole process cannot achieve end-to-end training. In [34], bi-temporal RS images were combined into one image as input to the modified U-Net [35] network structure. In [36], the improved semantic segmentation network UNet++ [37] was introduced for application to the change detection task. Moreover, in order to obtain global information and fine boundary information, deeply supervised image fusion strategies are used in the encoder structure, which significantly improves the CD of VHR RS images. In [11], a Siamese fully convolutional network structure was proposed to extract the feature information of image pairs by using two identical network structures in the same manner as shared weights. However, these methods do not focus on the extremely unbalanced samples in the CD task and the rules of learning the features of the deep learning model.

### B. Attention Mechanism

The attention mechanism is an important module in machine learning, and is widely used in various types of machine learning tasks, including natural language processing [12], image recognition [38], and speech recognition [39]. The attention mechanism helps the model to assign different weights to each part of the input and extract more critical and important information. It can ensure that the model makes a more accurate judgment, and at the same time, does not introduce any additional calculation and storage overhead. Similarly, the attention mechanism has been introduced into each of the RS CD tasks listed above. In [40], spatial attention modules (SAM) and channel attention modules (CAM) were used to improve the boundary integrity and semantic consistency of the change feature map. In [13], the performance of CD is improved by exploring the relationship between pixels in different channels and space, and the self-attention mechanism is used to calculate the weights of two pixels in different bi-temporal images to generate discriminative features. In [41], a model based on super-resolution and stacking attention is proposed to improve the CD performance. In [42], a method based on differential image guidance and an attention model is proposed to describe the interval correlation of low-level and high-level features. Although these methods can improve the details of semantic information in CD tasks, there is still considerable room for improvement and optimization.

### III. PROPOSED HANET

In this section, we first introduce the motivation behind the proposed method, then present the model details. Fig. 4. shows the overall architecture of our proposed HANet.

### A. Progressive Foreground-Balanced Sampling (PFBS)

We propose a Progressive Foreground-Balanced Sampling (PFBS) approach that enables the model to accurately learn the features of the foreground image during the early training process. The basic idea behind PFBS is to improve the influence of changed pixels on model training, and moreover, to solve the sample imbalance problem by training the foreground image first and then the background image slowly thereafter. For ease of understanding, we here use the WHU-CD dataset for an example presentation, which contains a total of 4536 training set images (including 1200 foreground images and 3336 background images). As visualized in Fig.3, there are four lines representing the four training processes of Normal Train, Fixed-X, Linear-Y and Fixed-X Linear-Y. These are introduced in more detail below.

> ➢ **Normal Train** means that all training datasets are used throughout the whole training process. Therefore, the model learns the entire dataset at each epoch, completely ignoring the fact that this binary dataset is extremely unbalanced.

> ➢ **Fixed-X** means that the first X epochs are trained with 1200 foreground images, after which the whole training dataset is used for training. Notably, Fixed-X takes the data imbalance into consideration. It first helps the model to learn the "change" features by specifically learning the foreground images with changing information, then enhances the robustness of the model by learning the entire dataset (including background images).

> ➢ **Linear-Y** means that the first Y epochs are trained by gradually and linearly increasing the ratio of background images to foreground images, after which the whole training dataset is used for training. Similarly, Linear-Y also takes the imbalance of data into consideration. It only learns the features of the foreground images at the time of the first epoch, the gradually learns the features of the background images at the time of the second epoch to epoch Y, and then learns the features of the entire dataset after epoch Y+1. In this example, 222 background images are added per epoch.

> ➢ **Fixed-X Linear-Y** refers to the combination of **Fixed-X** and **Linear-Y**. Under this approach, the model learns the foreground images at the first X epoch, then adds the background images linearly through Y epochs, and subsequently learns the entire dataset after X+Y epochs. In this example, 333 background images are added per epoch.

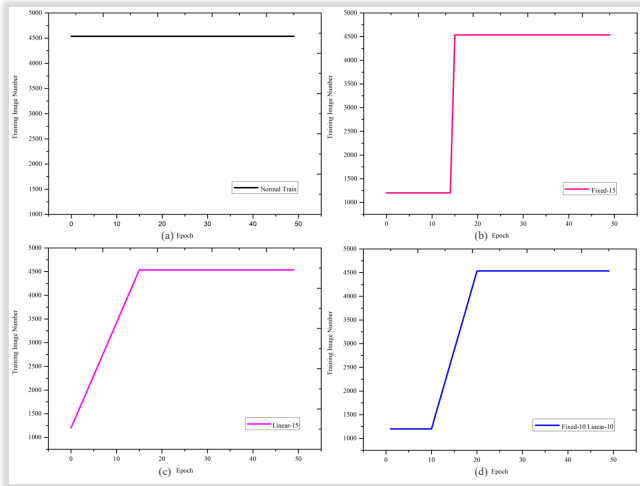

**Fig. 3.** Visualization of the proposed Progressive Foreground-Balanced Sampling (PFBS) process.

*B. HANet details*

As shown in Fig.4, HANet is a Siamese architecture that incorporates a weight-shared feature extractor. It consists of two parts: a multi-scale feature extractor and the HAN module. The multi-scale feature extractor can obtain multi-scale building semantic information. For its part, the HAN module can gradually obtain semantic integration and refinement features of buildings. The deep learning-based CD method generally takes two different temporal images (T1 and T2) as input images, and produces one prediction image as the output image. The process is described in more detail below.

Our HANet model includes a convolutional block, adaptive average pooling layer, and HAN module, which refer the MSPSNet [15]. T1 and T2 are of the same size; the dimension of T1 is $W \times H \times C$, where $W$ is the width of the input image, $H$ is the height of the input image, and $C$ is the number of input image channels. For the VHR images used as input pairs, we use the size $256 \times 256 \times 3$. In our HANet, we extract four-scale building features in four steps. Every scale feature can be described by $\{f_i^m | i = 1,2, m = 1,2,3,4\}$. Here, $f_i^m$ means each temporal image, $i$ represents T1 and T2, and $m$ denotes the four different scales. When we get the first $f_i^m$, it will simultaneously be taken as the input of the HAN module. The HAN module can continuously help the model to identify the changed regional features and improve the feature details.

In general, our HANet architecture consists of four convolutional blocks and three adaptive average pooling layers. In a convolutional block, a residual connection is used to superimpose the feature. The order is as follows: the convolutional layer, batch normalization (BN) layer, and activation functions rectified linear unit (ReLU). The convolutional layer, BN, and ReLU repeat twice. Therefore, the result of the convolutional block can be denoted as follows:

$$g(f_i^m) = R(B_N(conv_1^{3\times3}(f_i^m))) \quad (1)$$
$$Output_0 = R(B_N((conv_2^{1\times1}(g(f_i^m))) + conv_1^{3\times3}(f_i^m)) \quad (2)$$

where $\{f_i^m | i = 1,2, m = 1,2,3,4\}$ represents the input features. $conv_1^{3\times3}$ represents the first convolutional layer with a kernel size of 3. $conv_2^{1\times1}$ represents the second convolutional layer with a kernel size of 1. $B_N$ is the normalization operation, and $R$ is the activation operation.

In the adaptive average pooling layer, we use 128-,64-, and 32-layer features to obtain the multi-scale feature blocks. Thus, our HANet can easily learn how to gradually and deeply extract semantic features, which makes it discriminative.

*C. HAN module*

As shown in Fig.5, our HAN module consists of a parallel convolution structure (PCS) [15] and spatial-spectral axial attention, and the latter contains both Column and Row Attention (Col.-Row-A) [43], [44]. PCS can increase the effectiveness of feature extraction and reduce the model computational cost. The calculation amount of self-attention is second order, and Col.-Row-A can reduce the calculation amount and achieve higher calculation efficiency. The input first passes through PCS for multi-scale feature fusion, then passes through the refinement of CAM and Col.-Row-A at the same time. Finally, the results of CAM and Col.-Row-A are added together to produce the results of the HAN module.

PCS has four different group convolutions [45] with a kernel size of $3 \times 3$ and four dilation parameters, enabling it to integrate multi-scale features. Group convolution can expand the receptive fields. $c$ is the number of group members in each group convolution operation, and is 0.5 times the size of the input data. Therefore, the PCS can be denoted as follows:

$$F = cat(f_1^m, f_2^m), m = 1,2,3,4 \quad (3)$$
$$S = cat(conv_{d=1}^{3\times3}(F), conv_{d=2}^{3\times3}(F),$$
$$conv_{d=3}^{3\times3}(F), conv_{d=4}^{3\times3}(F)) \quad (4)$$

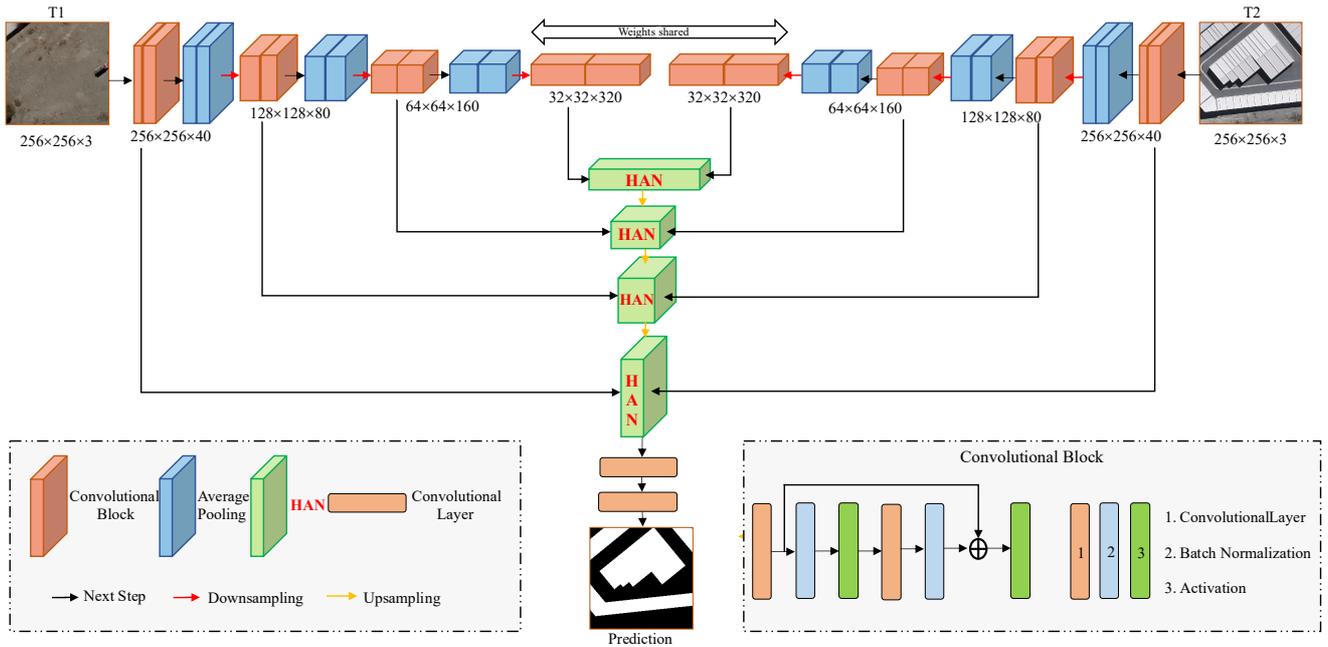

**Fig. 4.** Illustration of our proposed HANet model.

$$Output_1 = conv^{1\times1}(S) \qquad (5)$$

where $cat(\cdot)$ represents the channel dimension concatenation operation, $conv_{d=m}^{3\times3}$ is the convolutional layer with kernel size 3 and dilation of $m$, and $conv^{1\times1}$ represents the $1\times1$ convolutional layer.

CAM[46], [47] is the abbreviation of the channel attention module, which can refine the detailed feature. It contains a $3\times3$ and a $1\times1$ convolutional layer, an element-wise sum operation, two matrix multiplication operations, a transpose operation, three reshape operations, and an activation function. Therefore, the CAM can be denoted as follows:

$$I_0 = conv_1^{3\times3}(Output_1) \qquad (6)$$

$$Output_2 = conv_2^{1\times1}\left(S_o(R_e(I_0) \times R_e(T_r(I_0)))\right.$$
$$\left. \times R_e(I_0) + I_0\right) \qquad (7)$$

where $conv_1^{3\times3}$ is the initiatory convolutional layer with a kernel size of 3, while $R_e$ represents the reshape operation. $T_r$ represents the transpose operation. $S_o$ is the SoftMax function, and $conv_2^{1\times1}$ is the last convolutional layer with a kernel size of 1.

Col.-Row-A is very similar to CAM in implementation, with the major difference relating to the axial attention. The amount of computation for CAM is second-order, and Col.-Row-A can reduce the amount of computation required and achieve higher computational efficiency. Col.-Row-A is calculated first in the horizontal direction and then in the vertical direction to reduce the computational complexity. The receptive field of Col.-Row-A is $W$ (or $H$) pixels in the same row (or column) of the target pixel, meaning that it has a much smaller receptive field than CAM. Therefore, the column attention can be denoted as follows:

$$I_1 = conv_1^{3\times3}(Output_1) \qquad (8)$$

$$Output_3 = conv_2^{1\times1}\left(S_o(R_e^c(I_1) \times R_e^c(T_r(I_1)))\right.$$
$$\left. \times R_e^c(I_1) + I_1\right) \qquad (9)$$

Here, $R_e^c$ represents the reshape operation of column attention. The row attention can be denoted as follows:

$$I_2 = conv_1^{3\times3}(Output_3) \qquad (10)$$
$$Output_4 = conv_2^{1\times1}(S_o(R_e^r(I_2) \times R_e^r(T_r(I_2))) \times$$
$$R_e^r(I_2) + I_2) \qquad (11)$$

Here, $R_e^r$ represents the reshape operation of row attention. Finally, we can create HAN module by adding CAM and Col.-Row-A.

$$Output_5 = Output_2 + Output_4 \qquad (12)$$

Through the continuous improvement with PCS, CAM and Col.-Row-A, context information, the interior, and the edge features of the building can be better extracted. In fact, our HAN module considers not only multi-scale information extraction, but also contextual information extraction, as well as the number of modules parameter.

*D. Loss Function*

For this extremely unbalanced change detection challenge, we adopt hybrid loss [48] as the loss function to alleviate the impact of the data imbalance. Hybrid loss combines weighted cross-entropy loss and dice loss, and can be expressed as follows:

$$L = L_w + L_d \quad (13)$$

$$L_w = \frac{1}{W \times H}\sum_{i=1}^{W \times H} w[cla] \cdot \left(\log\left(\frac{exp(\hat{y}[i][cla])}{\sum_{l=0}^{1} exp(\hat{y}[i][l])}\right)\right) \quad (14)$$

$$L_d = 1 - \frac{2 \cdot S_o(\hat{Y})}{Y + S_o(\hat{Y})} \quad (15)$$

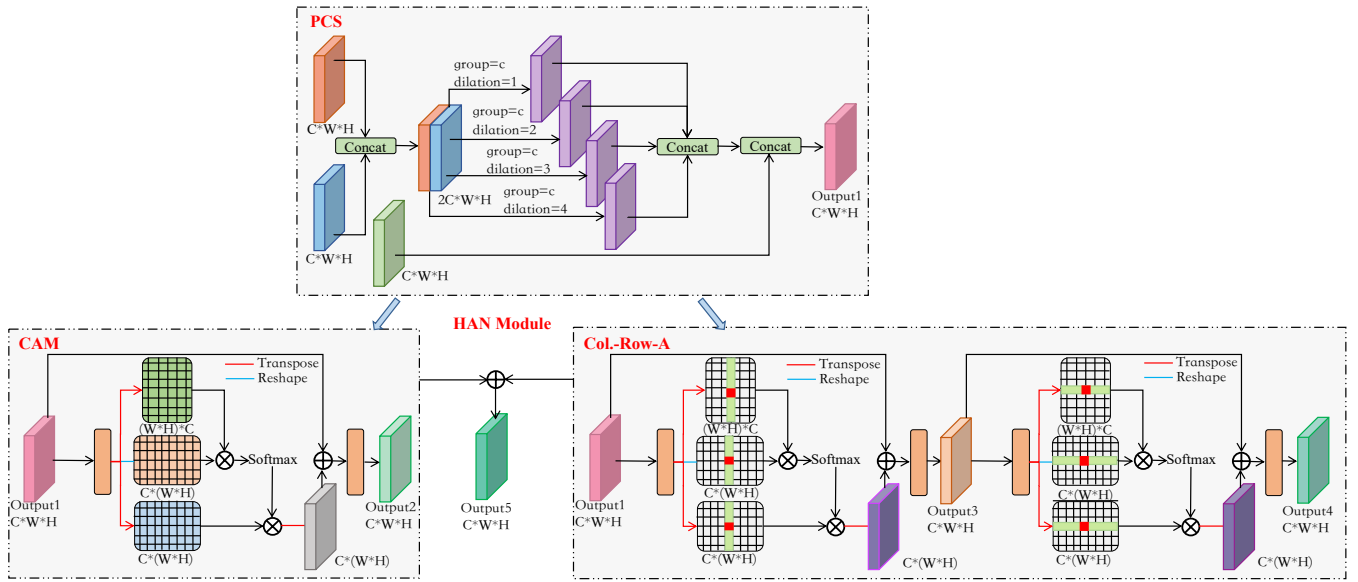

**Fig.5.** Illustration of our HAN module. Here, HAN means Hierarchical Attention Network, PCS means Parallel Convolutional Structure, CAM means Channel Attention Module, and Col.-Row-A means Column Attention and Row Attention

Here, $L_w$ represents the weighted cross-entropy loss, $L_d$ represents the dice loss, $w$ represents the weights, the value of $cla$ is either 1 or 0 (corresponding to changed and unchanged pixels respectively), $\hat{y}[i]$ represents the $i$ th point in $\hat{Y}$, $i$ and $l$ are indexes, $\hat{Y} = \{\hat{y}[i], i = 1,2,\dots,W \times H\}$ represents the change map, and $\hat{Y}$ represents the ground truth.

## IV. EXPERIMENT

In this section, we introduce the experimental datasets and environment, comparison models, and evaluation metrics. We then discuss the experimental results and ablation study in detail.

### A. Experimental setup

**WHU-CD**[21] is a public remote sensing Building Change Detection dataset, which contains one large VHR (0.2m/pixel) image patch pair with a size of 32,507 ×15,345, which is cropped by 512 × 512. It includes areas of Christchurch in New Zealand, where a 6.3-magnitude earthquake occurred in 2011, after which significant rebuilding took place. There are 21,442,501 changed pixels, accounting for 4.26% of the total, and 481,873,979 unchanged pixels, accounting for 95.74%. It is therefore an extremely unbalanced binary classification dataset (the changed pixels in the label are 1, and the unchanged pixels are 0), as is shown in Fig.2. We adopted the default data split (training: 1260 image pairs; testing: 690 image pairs) published on the authors' websites. Due to GPU memory capacity limitations, and to facilitate a fair comparison with other algorithms, we directly cropped the default image patch pairs into sizes of 256 × 256 with no overlap. Also, we randomly selected 10% of images from the training dataset to form the validation dataset. Therefore, we obtained a dataset including 4536/504/2760 pairs of patches for training/validation/testing respectively.

**LEVIR-CD** [13] is a public large-scale remote sensing Building Change Detection dataset, which consists of 637 VHR (0.5m/pixel) image patch pairs with a size of 1024 × 1024 pixels. It includes 20 different areas in several cities in the United States, and contains a large number of seasonal and light changes, which makes change detection more difficult. There are 31,066,643 changed pixels, accounting for 4.65% of the total, and 636,876,269 unchanged pixels, accounting for 95.35%. It is therefore an extremely unbalanced binary classification dataset (the changed pixels in the label are 1, and the unchanged pixels are 0), as shown in Fig.2. We adopted the default data split (training: 445 image pairs; validation: 64 image pairs; testing: 128 image pairs) published on the authors' websites. Due to GPU memory capacity limitations, and to facilitate a fair comparison with other algorithms, we directly cropped the default image patch pairs into sizes of 256 × 256 with no overlap. Therefore, we obtained a dataset including 7120/1024/2048 pairs of patches for training/validation/testing respectively.

**Implementation details**. Our models are implemented on PyTorch and trained using a single NVIDIA RTX 3090 GPU. We adopt the Adam optimizer with a weight decay 5e-4 and learning rate 5e-4 (with the gamma of 0.5 adopted to update the learning rate) to minimize the loss. We use StepLR with a step size of 8 and gamma of 0.5 to update our learning rate, as shown in Equation 16. Due to the limitations of the GPU, we set a batch size of 8 and epoch number of 100 to make the model converge. We trained our model for 100 epochs and saved the best model on the validation set as the final training result. We refer to the fixed foreground image training method as **Fixed-X** and the linear foreground image increase method as **Linear-X**; here, X means the number of epochs.

$$New_{lr} = initial_{lr} \times \gamma^{epoch//step\ size} \quad (16)$$

**Evaluation Metrics**. In order to quantitatively verify the effectiveness of our proposed model, we use the following metrics for evaluation: F1-score ($F1$), Precision ($Pre.$), Recall ($Rec.$), Overall Accuracy ($OA$), and Kappa Coefficient ($KC$). These are employed by comparing the GT and prediction maps, and can be specifically defined as follows:

$$F1 = \frac{2}{Pre.^{-1} + Rec.^{-1}} \quad (17)$$

$$Pre. = TP/(TP + FP) \quad (18)$$

$$Rec. = TP/(TP + FN) \quad (19)$$

$$OA = (TP + TN)/(TP + TN + FN + FP) \quad (20)$$

$$KC = \frac{OA - PRE}{1 - PRE} \quad (21)$$

$$IoU = TP/(TP + FN + FP) \quad (22)$$

$$PRE = \frac{(TP+FN)\times(TP+FP)}{(TP+TN+FP+FN)^2} + \frac{(TN+FP)\times(TN+FN)}{(TP+TN+FP+FN)^2} \quad (23)$$

where TP denotes the number of true positives, TN denotes the number of true negatives, FP denotes the number of false positives, and FN denotes the number of false negatives. "PRE" represents the sum of the "ground truth and the product of the predicted result" corresponding to all categories, divided by the "average of the total sample data".

It is worth noting that higher values of F1, OA, and KC indicate good CD performance.

*B. Comparison with state-of-the-art methods*

It is necessary to compare our proposed model with three different types of state-of-the-art methods (pure CNN-based methods, attention-based methods, and transformer-based methods). Accordingly, we choose three classic pure CNN-based models (FC-EF [11], FC-Siam-conc [11], and FC-Siam-diff[11]), three attention-based models (STANet [13], SNUNet [14], and MSPSNet [15]), and three transformer-based models (BIT [17], Change Former [18], and RSP-BIT [19]).

➢ **FC-EF** [11]: A Fully Convolutional Early Fusion network, which is directly based on the U-Net structure, that concatenates two input images before feeding them into the network, then treats them as different channels of an image.

➢ **FC-Siam-conc** [11]: A Fully Convolutional Siamese-Concatenation model, which skip-connects the three feature diagrams from the two encoder branches and the corresponding layer of the decoder.

➢ **FC-Siam-diff** [11]: A Fully Convolutional Siamese-Difference model, which first obtains the absolute value of the difference between the two decoder branches, after which a skip connection is made with the corresponding layer of the decoder.

➢ **STANet** [13]: A Siamese-based spatial-temporal attention neural network, which can get more discriminative features through use of the Spatial-Temporal Attention Module.

➢ **SNUNet** [14]: A densely connected Siamese network, which alleviates the loss of localization information in the deep layers of neural networks and refines the most representative features of different semantic levels via ECAM (Ensemble Channel Attention Module).

➢ **MSPSNet** [15]: A deep multiscale Siamese network, which obtains the multiscale feature via PCS (Parallel Convolutional Structure) and SA (Self-Attention).

➢ **BIT** [17]: A bitemporal image transformer network, which can efficiently and effectively model contexts within the spatial-temporal domain by using the transformer to capture the contextual information between different temporal images.

➢ **Change Former** [18]: A transformer-based Siamese network architecture, which unifies a hierarchically structured transformer encoder with a Multi-Layer Perception (MLP) decoder in its Siamese network architecture to efficiently render multi-scale long-range details.

- **RSP-BIT**[19]: A BIT model with RSP (Remote Sensing Pretraining), which adopts the ViTAE [49] Network and uses the MillionAID [50] dataset for pretraining.

Notably, we implement these CD models using their public codes with default hyperparameters found on GitHub in the same software environment.

Tab. I shows the specific quantitative comparison results on the WHU-CD and LEVIR-CD datasets obtained by these methods. It can be observed that our HAN module is a lightweight and effective self-attention mechanism by calculating the parameters and floating-point operations (FLOPs) of the model. By observing the data in red, we can see that our proposed HANet achieves good performance. Moreover, on the key metrics of F1-score, OA, and KC, HANet also achieves better performance. In particular, the WHU-CD and LEVIR-CD datasets are among the more unbalanced samples from Fig.2. For example, the F1-score of our HANet exceeds that of Change Former by 0.98/0.08 points on these two datasets respectively. Getting such a result is a little difficult because Change Former is a transformer-based method, which has more network parameters. Note that our HANet is better at integrating contextual semantic information at different scales. This may be attributed to the powerful semantic information extraction ability of our HAN module.

Fig.6. presents the visualization comparison results on the WHU-CD and LEVIR-CD datasets with these methods. We selected some challenging samples (a)–(h) with complex building features from the test set for illustrative purposes. This diagram can be used to intuitively compare the performance of each model. To further improve the intuitiveness of this visualization, we use several colors and a small red box to show the key and detailed results of the model. Blue and red colors indicate missed detection and error detection respectively.

First, our HANet is better able to avoid false negatives (e.g. Fig.6 (a), (b), (g), (h)) on complex building features. For example, the pure CNN-based FC-EF and FC-Siam-conc methods have obviously missed detections. Compared with the attention-based SNUNet and MSPSNet on (d), it can be easily seen that our proposed HANet has fewer blue pixels.

Moreover, when compared with the transformer-based methods BIT,Change Former and RSP-BIT in Fig.6 (c), HANet also demonstrates good performance when extracting small building semantic features.

Second, our HANet is better at avoiding false positives (e.g. Fig.6 (b),(d),(g)) on complex background features, including the irrelevant changes caused by seasonal variation and land use cover change. By observing the red area next to the building, we can determine that FC-EF, STANet, SNUNet, and MSPSNet are more sensitive to the changes in certain trees, which are not buildings. In Fig.6 (d), the U-shaped building undergoes a change in the impermeable material surface; our HANet is much better at extracting semantic changes in the region of interest.

Third, our HANet is better at extracting detailed building features (e.g. Fig. 6 (h)). In some non-conventional quadrilateral buildings, even if the building size is very large, our HANet is still better at extracting the edge features of irregular buildings.

### C. Ablation Study

We set up the following models to validate the effectiveness of our proposed model:
- **Base**: The baseline model consists of the CNN Siamese Network with PCS and CAM.
- **HANet**: Base model + Row-Col.-A.
- **HANet-Fixed-X**: HANet + Fixed-X.
- **HANet-Linear-X**: HANet + Linear-X.
- **HANet-Fixed-X Linear-Y**: HANet + Fixed X and then Linear-Y.

There are several hyper-parameters in PFBS and attention modules in the HAN module. Therefore, we choose some of them for an ablation study. We select some of the typical ablation study results for presentation in order to intuitively observe the performance. Here, the white area indicates prediction. Special attention should be paid to the four-color box: here, each color stands for a different kind of typical ablation study.

**Ablation on training stability.** Visualizing the training process of general CD method and HANet helps us to improve our understanding of the training process. From Fig.8, we can observe that the general CD method has an unstable process during convergence, around 20K iterations (equal to 20 epochs). After adding PFBS, HANet convergence becomes smooth, which means PFBS has a

TABLE I

RESULTS OF COMPARISON WITH OTHER SOTA CHANGE DETECTION METHODS IN TERMS OF PARAMETERS (PARA.), FLOPs, F1-SCORE, PRECISION, RECALL, OVERALL ACCURACY, AND INTERSECTION OVER UNION ON THE WHU-CD AND LEVIR-CD DATASETS. ALL VALUES ARE IN %. HIGHER VALUES OF F1 AND OA INDICATE GOOD CD PERFORMANCE. FOR CONVENIENCE: BEST, 2ND-BEST, AND 3RD-BEST.

| Model | Para. (M) | FLOPs(G) | WHU-CD | | | | | LEVIR-CD | | | | |
|---|---|---|---|---|---|---|---|---|---|---|---|---|
| | | | F1 | Pre. | Rec. | OA | IoU | F1 | Pre. | Rec. | OA | IoU |
| FC-EF[11] | 1.35 | - | 58.05 | 76.49 | 46.77 | - | 40.89 | 61.52 | 73.31 | 53.00 | - | 44.43 |
| FC-Siam-conc[11] | 1.54 | 2.29 | 63.99 | 72.06 | 57.55 | - | 47.05 | 64.41 | 95.30 | 48.65 | - | 47.51 |
| FC-Siam-diff[11] | 1.35 | 2.29 | 86.31 | 89.63 | 83.22 | - | 75.91 | 89.00 | 91.76 | 86.40 | - | 80.18 |
| STANet-PAM[13] | 16.93 | 6.58 | 82.00 | 75.70 | 89.30 | 98.60 | 69.44 | 85.20 | 80.80 | 90.10 | 98.40 | 74.22 |
| SNUNet[14] | 12.03 | 27.44 | 87.76 | 87.84 | 87.68 | 99.13 | 78.19 | 89.97 | 91.31 | 88.67 | 98.99 | 81.77 |
| MSPSNet[15] | 2.21 | 14.17 | 86.49 | 87.84 | 85.17 | 99.05 | 76.19 | 89.67 | 90.75 | 88.61 | 98.96 | 81.27 |
| BIT[17] | 3.55 | 4.35 | 80.97 | 74.01 | 89.37 | 98.51 | 68.02 | 89.94 | 90.33 | 89.56 | 98.98 | 81.72 |
| ChangeFormer[18] | 20.75 | - | 87.18 | 92.70 | 82.28 | 99.14 | 77.27 | 90.20 | 92.05 | 88.37 | 99.01 | 82.21 |
| RSP-BIT[19] | 24.44 | - | 78.50 | 69.93 | 89.45 | 98.26 | 64.60 | 89.71 | 92.00 | 87.53 | 98.98 | 81.34 |
| HANet(ours) | 3.03 | 14.07 | 88.16 | 88.30 | 88.01 | 99.16 | 78.82 | 90.28 | 91.21 | 89.36 | 99.02 | 82.27 |

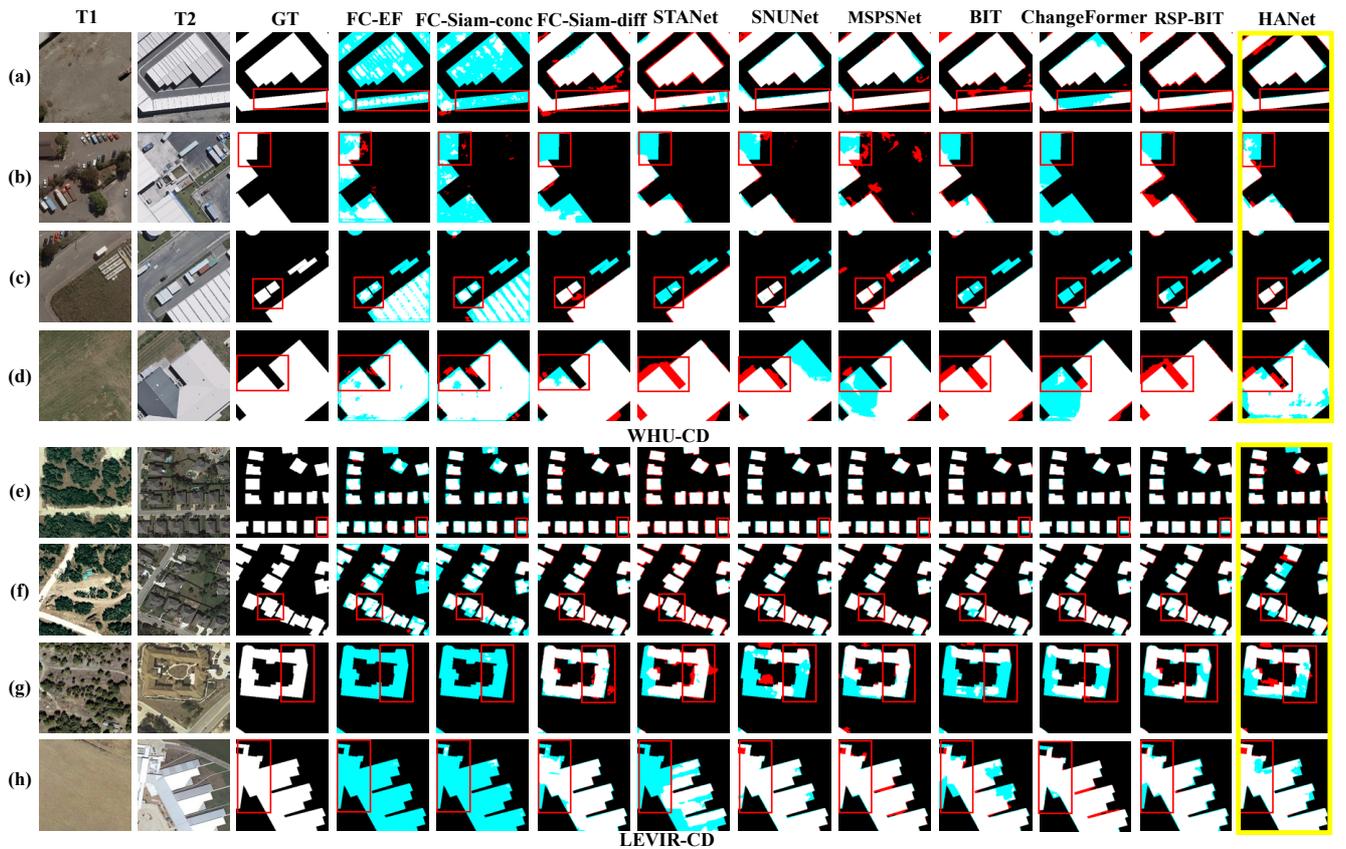

**Fig. 6.** Visualization comparison of different methods on the WHU-CD and LEVIR-CD test sets. For convenience, several colors are used to facilitate a clearer visualization of results: i.e., TP (true positive, white), FP (false positive, red), TN (true negative, black), and FN (false negative, blue).

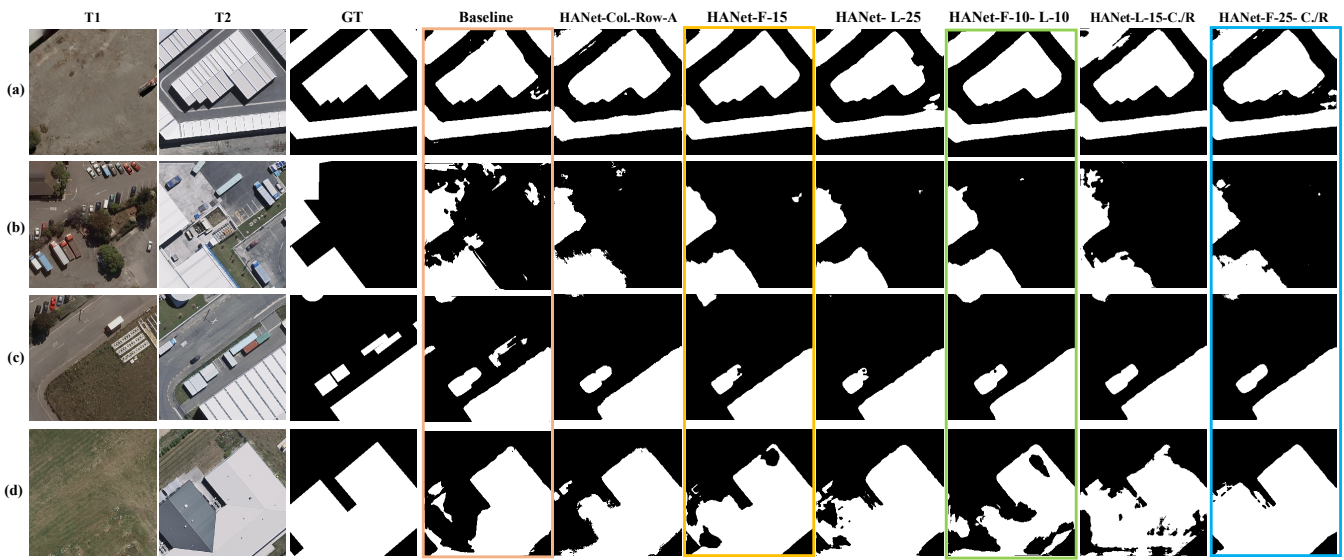

**Fig. 7.** V*isualization of ablation study results*.

good effect. Under these circumstances, HANet finds it easier to converge.

**Ablation on PFBS.** In the Fixed-X of PFBS, we discussed the hyper-parameter of epoch numbers. As shown in Tab. II, we conducted an experiment at intervals of 5 epochs from 5 to 40 epochs. We present the higher results of the 5, 15 and 25 epochs. Regardless of which parameter we use, the results exceed the baseline. Moreover, the training result of Fixed-15 reached 88.49% in terms of the F1-score, which is a parameter with the best performance.

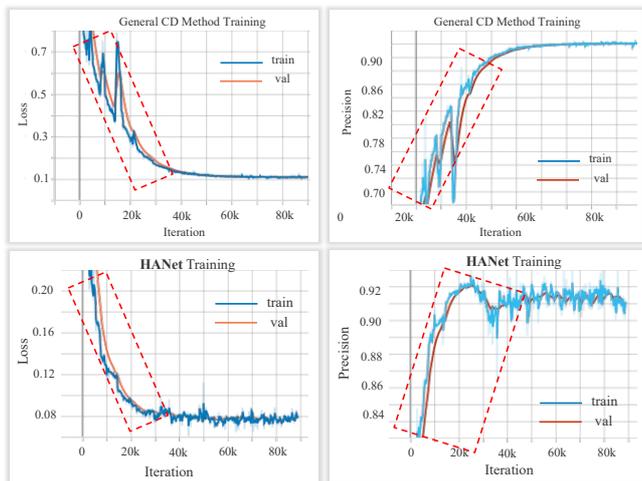

**Fig. 8.** The visualization of general CD method's and HANet's training process with regard to loss and precision.

As a result, as shown in Fig.7, Fixed-15 also achieves better performance than the baseline. This shows that the building semantic features in the foreground images of WHU-CD can be learned in 15 epochs. This result shows that different datasets should have different standards in this PFBS.

TABLE II

ABLATION STUDY ON THE FIXED-X (F-X) OF PFBS ON WHU-CD. F1-SCORE, PRECISION, AND RECALL ARE COMPARED.

| Model | Fixed-X on WHU-CD | | | | F1 | Pre. | Rec. |
|---|---|---|---|---|---|---|---|
| | Base | +F-5 | +F-15 | +F-25 | | | |
| Base | √ | | | | 84.36 | 90.44 | 81.03 |
| +F-5 | | √ | | | 86.68 | 85.45 | 87.96 |
| **+F-15** | | | √ | | **88.49** | **88.99** | **88.00** |
| +F-25 | | | | √ | 88.27 | 90.25 | 86.37 |

TABLE III

ABLATION STUDY ON THE FIXED-X (F-X) AND LINEAR-X (L-X) OF PFBS ON WHU-CD. F1-SCORE, PRECISION, AND RECALL ARE COMPARED.

| Model | Fixed-X + Linear-X on WHU-CD | | | | F1 | Pre. |
|---|---|---|---|---|---|---|
| | Base | +L-25 | +F-10 L-10 | +F-15 L-15 | | |
| Base | √ | | | | 84.36 | 90.44 |
| +L-25 | | √ | | | 87.18 | 89.79 |
| **+F-10 L-10** | | | √ | | **88.10** | **88.78** |
| +F-15 L-15 | | | | √ | 86.41 | 86.82 |

Similarly, we discuss the hyper-parameters of Linear-X and the combination of Fixed-X and Linear-Y. As shown in Tab. III, regardless of which parameters are employed, the results of our approach can exceed the baseline. Note that there are many possible combinations of X and Y. Experiments have shown that the use of more foreground images for training does not necessarily produce better performance; in fact, the performance depends on the scale of the foreground image. As shown in Fig.7, Fixed-10 plus Linear-10 yields relatively good results.

TABLE IV

ABLATION STUDY ON THE KEY PART OF HAN MODULE (COL.-ROW-A, ABBREVIATED TO C./R) ON WHU-CD. F1-SCORE, PRECISION, AND RECALL ARE COMPARED.

| Model | Row/Col.-Attention on WHU-CD ||||  |  |  |
|---|---|---|---|---|---|---|---|
|  | Base | +Col. | +Row | +C./R | F1 | Pre. | Rec. |
| Base | √ |  |  |  | 84.36 | 90.44 | 81.03 |
| +Col. |  | √ |  |  | 87.31 | 87.84 | 86.78 |
| +Row |  |  | √ |  | 87.73 | 88.29 | 87.18 |
| +C./R |  |  |  | √ | **87.78** | **87.45** | **88.12** |

**Ablation on HAN module.** In the HAN module, the key component is attention, which enables contextual semantic information about the building to be obtained. As shown in Tab. IV, we investigate the performance of axial attention, including column attention and row attention. It is obvious that single-column attention or row attention produces different scores, because the model initializes parameters randomly. Experiments show that the combination of column and row attention is superior to single axial attention.

**Ablation on PFBS and HAN module.** We next discuss the combination of our proposed PFBS and HAN module. As shown in Tab. V, the performance of both Linear-X and Fixed-X plus HAN module exceed the baseline. Moreover, the performance of Fixed-X plus HAN module is better than that of Linear-X plus HAN module. From Fig.7, we can observe that Fixed 25 plus Col.-Row-A is better than Linear-15 plus Col.-Row-A.

TABLE V

ABLATION STUDY ON THE FIXED-X (F-X) AND LINEAR-X (L-X) OF PFBS AND THE KEY PART OF THE HAN MODULE (COL.-ROW-A, ABBREVIATED TO C./R) ON WHU-CD. F1-SCORE, PRECISION, AND RECALL ARE COMPARED.

| Model | Row/Col.-A-Fixed-X + Linear-X on WHU-CD |||||||
|---|---|---|---|---|---|---|---|
|  | Base | +L-15 C./R | +L-25 C./R | +F-15 C./R | F1 | Pre. | Rec. |
| Base | √ |  |  |  | 84.36 | 90.44 | 81.03 |
| +L-15 C./R |  | √ |  |  | 87.57 | 87.35 | 87.78 |
| +L-25 C./R |  |  | √ |  | 87.26 | 89.15 | 85.45 |
| **+F-25 C./R** |  |  |  | √ | **88.16** | **88.30** | **88.01** |

**Ablation of PFBS on another method.** We add our proposed PFBS into MSPSNet [15] to validate the efficiency of PFBS. As shown in Tab. VI, the three methods of PFBS (including Fixed-X, Linear-X, and the combination of Fixed-X and Linear-X) all work well on MSPSNet. Moreover, the performance of the combination of Fixed-X and Linear-X is best. Therefore, we conclude that it is possible to use PFBS to improve the performance of another method.

TABLE VI

ABLATION STUDY OF PFBS ON MSPSNET.

| Model | WHU-CD ||||||
|---|---|---|---|---|---|---|
|  | F1 | Pre. | Rec. | OA | KC | IoU |
| MSPSNet [15] | 86.49 | 87.84 | 85.17 | 99.05 | 86.00 | 76.19 |
| +F-10 | 87.50 | 87.66 | 87.34 | 99.11 | 87.04 | 77.77 |
| +L-15 | 86.88 | 87.63 | 86.14 | 99.08 | 86.40 | 76.80 |
| **+F-5 L-10** | **87.65** | **87.25** | **88.05** | **99.12** | **87.19** | **78.01** |

TABLE VII

ABLATION STUDY OF THE MODEL WITH DIFFERENT LOSS FUNCTIONS INCLUDING HYBRID LOSS(HL) AND FOCAL LOSS(FL) ON WHU-CD.

| Loss Model | Different loss functions on WHU-CD ||||| 
|---|---|---|---|---|---|
|  | HL | FL | HL+ F-15 | FL+ F-15 | F1 |
| Hybrid Loss | √ |  |  |  | 84.36 |
| Focal Loss |  | √ |  |  | 85.50 |
| Hybrid Loss+ F-15 |  |  | √ |  | **88.49** |
| Focal Loss+ F-15 |  |  |  | √ | 87.38 |

**Ablation of different loss functions.** In order to verify the effectiveness of PFBS, we compare the performance of different loss functions on WHU-CD. In this paper, we use Focal Loss[51] and Hybrid Loss to compare. As shown in Tab. VII, we can know that the performance of Focal Loss is better than Hybrid Loss. Subsequently, our proposed PFBS (here we use Fixed-15) simultaneously improve the F1-score and whereas Hybrid Loss with Fixed-15 is better.

TABLE VIII
ABLATION STUDY OF SAMPLING STRATEGIES ON LEVIR-CD.

| Sampling strategies | LEVIR-CD | | |
|---|---|---|---|
| | **100%** | 20% | 5% |
| | F1 | | |
| RSP-BIT [19] | 89.71 | 74.73 | 55.62 |
| SaDL [24] | 88.74 | 87.25 | **79.44** |
| CDNet+IAug [23] | 89.00 | **87.50** | 76.00 |
| HANet-PFBS(F-15) | **90.28** | 81.36 | 68.66 |

**Ablation of different sampling strategies.** There are some existing methods to address the class imbalance in CD except for using the loss function. For example, synthesise CD samples[23] that are more class-balanced, or transfer a pre-trained model[24] that is more robust. RSP-BIT[19] is a change detection method for remote sensing pretraining (RSP) on the MillionAID dataset. The following TABLE shows that our performance is best when we use 100% of the training dataset. And SaDL and CDNet +IAug are better when we use 5% and 20% of the training dataset. We further analyze the reason and find that their method increases the number of pixels with change information, while our PFBS method improves the performance of the model without increasing the number of pixels with change information.

## V. CONCLUSION

In this paper, a progressive foreground-balanced sampling (PFBS) approach on the basis of not adding change information is proposed to help the model accurately learn the features of the foreground image during the early stages of the training process. A discriminative Siamese network, HANet, is proposed to integrate multi-scale features and refine detailed features. Extensive experimental validation on two extremely unbalanced binary CD datasets (WHU-CD and LEVIR-CD) shows that our proposed methods (PFBS and HANet) outperform many remarkable existing models. In future work, it would be worthwhile to address some problems with the existing PFBS: for example, the optimal solution of X and Y in Fixed-X and Linear-Y, more forms of PFBS (such as the method of nonlinear increase), and the performance of PFBS on additional models. Similarly, the HAN module in HANet is suitable for migration to test performance on other models.

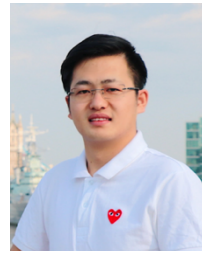

**Chengxi Han** (Student Member, IEEE) received the B.S. degree from the School of Geosciences and Info-Physics, Central South University, Changsha, China, in 2018. He is currently working toward a Ph. D. degree in photogrammetry and remote sensing at the State Key Laboratory of Information Engineering in Surveying, Mapping and Remote Sensing, Wuhan University, Wuhan, China.

His research interests include deep learning and remote sensing image change detection. He was a Trainee at the United Nations Satellite Centre (UNOSAT) of the United Nations Institute for Training and Research (UNITAR). He is the IEEE GRSS Wuhan Student Branch Chapter Chair since 2021. His chapter won IEEE GRSS 2022 Student Chapter Excellence Award.

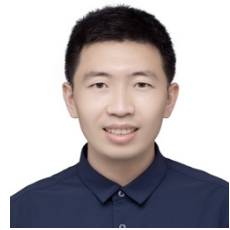

**Chen Wu** (M'16) received B.S. degree in surveying and mapping engineering from Southeast University, Nanjing, China, in 2010, and received the Ph.D. degree in Photogrammetry and Remote Sensing from State Key Lab of Information Engineering in Surveying, Mapping and Remote sensing, Wuhan University, Wuhan, China, in 2015.

He is currently a Professor with the State Key Laboratory of Information Engineering in Surveying, Mapping and Remote Sensing, Wuhan University, Wuhan, China. His research interests include multitemporal remote sensing image change detection and analysis in multispectral and hyperspectral images.

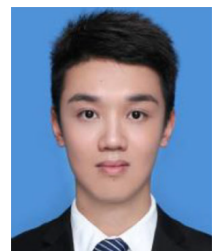

**Haonan Guo** received the B.S. degree in Sun Yat-sen University, Guangzhou, China, in 2020. He is currently working toward the Ph.D.degree with the State Key Laboratory of Information Engineering in Surveying, Mapping, and Remote Sensing,

Wuhan University, Wuhan, China.

His research interests include deep learning, building footprint extraction, urban remote sensing, and multisensor image processing.

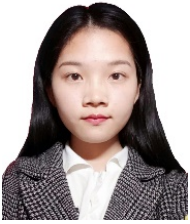

**Meiqi Hu** (Graduate Student Member, IEEE) received the B.S. degree in surveying and mapping engineering from the School of Geoscience and info-physics, Central South University, Changsha, China, in 2019. She is currently pursuing the Ph.D. degree with the State Key Laboratory of Information Engineering in Surveying, Mapping, and Remote sensing, Wuhan University, Wuhan, China.

Her research interests include deep learning and multitemporal remote sensing image change detection.

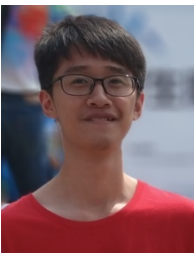

**Hongruixuan Chen** (Student Member, IEEE) received the B.E. degree in surveying and mapping engineering from the School of Resources and Environmental Engineering, Anhui University, Hefei, China, in 2019, and the M.E. degree in photogrammetry and remote sensing from State Key Laboratory of Information Engineering in Surveying, Mapping and Remote Sensing, Wuhan University, Wuhan, China, in 2022. He is now pursuing his Ph.D. degree at the Graduate School of Frontier Science, The University of Tokyo, Chiba, Japan. His current research fields include deep learning, domain adaptation, and multimodal remote sensing image interpretation and analysis. He was a Trainee at the United Nations Satellite Centre (UNOSAT) of the United Nations Institute for Training and Research (UNITAR). He also acts as a reviewer for eight international journals, e.g. IEEE TIP, TGRS, GRSL, and JSTARS.